\documentclass[runningheads]{llncs}
\usepackage{graphicx}
\usepackage{amsmath}
\usepackage{hyperref}
\usepackage{cleveref}
\usepackage{float}

\newcommand{\etal}{\textit{et al}. }

\begin{document}
\title{NoduleNet: Decoupled False Positive Reduction for Pulmonary Nodule Detection and Segmentation}
\titlerunning{NoduleNet}

% Commented for double blind peer review 
\author{Hao Tang\inst{1}$ ^,$ \inst{2} \and Chupeng Zhang\inst{2} \and
Xiaohui Xie\inst{1}}

% index{Tang, Hao}
% index{Zhang, Chupeng}
% index{Xie, Xiaohui}

%
\authorrunning{H. Tang et al.}
\institute{Department of Computer Science, Univeresity of California Irvine \\
\email{\{htang6, xhx\}@uci.edu} \and
Deep Voxel Inc.\\
\email{\{chupengz\}@deep-voxel.com}
}

% \author{}
% \institute{}

%
\maketitle              % typeset the header of the contribution
\begin{abstract}
Pulmonary nodule detection, false positive reduction and segmentation represent three of the most common tasks in the computer aided analysis of chest CT images. Methods have been proposed for each task with deep learning based methods heavily favored recently. However training deep learning models to solve each task separately may be sub-optimal - resource intensive and without the benefit of feature sharing. Here, we propose a new end-to-end 3D deep convolutional neural net (DCNN), called NoduleNet, to solve nodule detection, false positive reduction and nodule segmentation jointly in a multi-task fashion. To avoid friction between different tasks and encourage feature diversification, we incorporate two major design tricks: 1) decoupled feature maps for nodule detection and false positive reduction, and 2) a segmentation refinement subnet for increasing the precision of nodule segmentation. Extensive experiments on the large-scale LIDC dataset demonstrate that the multi-task training is highly beneficial, improving the nodule detection accuracy by 10.27\%, compared to the baseline model trained to only solve the nodule detection task. We also carry out systematic ablation studies to highlight contributions from each of the added components. Code is available at \url{ https://github.com/uci-cbcl/NoduleNet}.

\end{abstract}
\section{Introduction}
    Lung cancer has the highest incidence and mortality rates worldwide \cite{bray2018global}. Early diagnosis and treatment of pulmonary nodules can increase the survival rate of patients. Computed tomography (CT) has been widely used and proved effective for detecting pulmonary nodules. However, manually identifying nodules in CT scans is often time-consuming and tedious, because a radiologist needs to read the CT scans slice by slice, and a chest CT may contain over 200 slices. Accurate and precise nodule segmentation can provide more in-depth assessment of the shape, size and change rate of the nodule. When nodule is identified, a follow up scan in 3 - 12 months is usually required to assess its growth rate \cite{kalpathy2016comparison}. The growth of the lung tumor may be an indicator for malignancy, and an accurate nodule segmentation can be used for measuring the growth rate of the nodule. 

In recent years, deep convolutional neural network has emerged as a leading method for automatically detecting and segmenting pulmonary nodules and have achieved great success. State-of-the-art frameworks for nodule detection often ustilize the 3D region proposal network (RPN) \cite{ren2015faster} for nodule screening \cite{tang2018automated,zhu2017deeplung,setio2017validation,liao2019evaluate}, followed by a 3D classifier for false positive reduction \cite{ding2017accurate,tang2019end}. Although single stage detector has also been proposed in \cite{khosravan2018s4nd}, their hit criteria was different from what was more commonly adopted \cite{setio2017validation}. Moreover,
% cascaded classifiers have been widely used and proved great success in natural image applications \cite{li2015convolutional}. 
the refinement provided by the extra classifiers may correct some errors made by the detectors. In terms of nodule segmentation, U-Net \cite{ronneberger2015u} and V-Net \cite{milletari2016v} like structure is predominantly used \cite{wang2017central,wu2018joint,aresta2018iw}. 
% These methods reported results on different dataset with models trained and tested on different number of nodules, but they achieved similar accuracy. 
In practice, a computer aided diagnosis (CAD) system for pulmonary nodule detection and segmentation often consists of several independent subsystems, optimized separately. 

There are some limitations on handling each task completely independent. First, it is time-consuming and resource intensive to train several deep convolutional neural networks. Although each component is designed for different purposes, they share the common procedure of extracting feature representations that characterize pulmonary nodules. Second, the performance of the whole system may not be optimal, because separately training several systems prevents communication between each other and learning intrinsic feature representations. Intuitively, the segmentation mask of the nodule should provide a strong guide for the neural network to learn discriminative features, which may in turn improve the performance of nodule detection. 

Although multi-task learning (MTL) and feature sharing offer an attractive solution to combine different tasks, a naive implementation may cause other problems \cite{cheng2018revisiting}. First, because of the mismatched goals of localization and classification, it may be sub-optimal if these two tasks are performed using the same feautre map. Second, a large receptive field may integrate irrelevant information from other parts of the image, which may negatively affect and confuse the classification of nodules, especially small ones. \cite{cheng2018revisiting} decoupled localization and classification to address the problem in natural imaging. However, completely separating the two tasks without sharing any feature extraction backbone, still prevents cross-talk between two networks and may not be the most efficient. Therefore, a decoupled false positive reduction, that pools features from early scales of the feature extraction backbone, is proposed to address this problem, which allows learning both task-independent and task-dependant features.

% Although multi-task learning (MTL) and feature sharing offer an attractive solution to combine different tasks together, a naive implementation may cause other problems \cite{cheng2018revisiting}. First, nodule candidate screening and false positive reduction are essentially learning different features. Nodule candidate screening needs to learn translation covariant features in order to localize nodules, while false positive reduction needs to learn translation invariant features to reduce hard false positives. These two learning process may contradict with each other if localization and classification are performed using the same feature map. Second, a large receptive field is desirable for nodule candidate screening which, however, may be problematic for false positive reduction. A large receptive field is achieved by performing multiple pooling operations. It can incorporate information from the whole image that is beneficial to localize nodule candidates. However, feature information from other part of the image may negatively affect and confuse the classification of nodules, especially small nodules. As a result, the common strategy adopted in natural image detection \cite{ren2015faster,he2017mask} that performs localization and classification on the same feature map may not be optimal. A decoupled false positive reduction, that pools features from early scales of the feature extraction backbone, is proposed to address this problem.

\begin{figure}[!h]
\centering
 \includegraphics[width=0.9\textwidth]{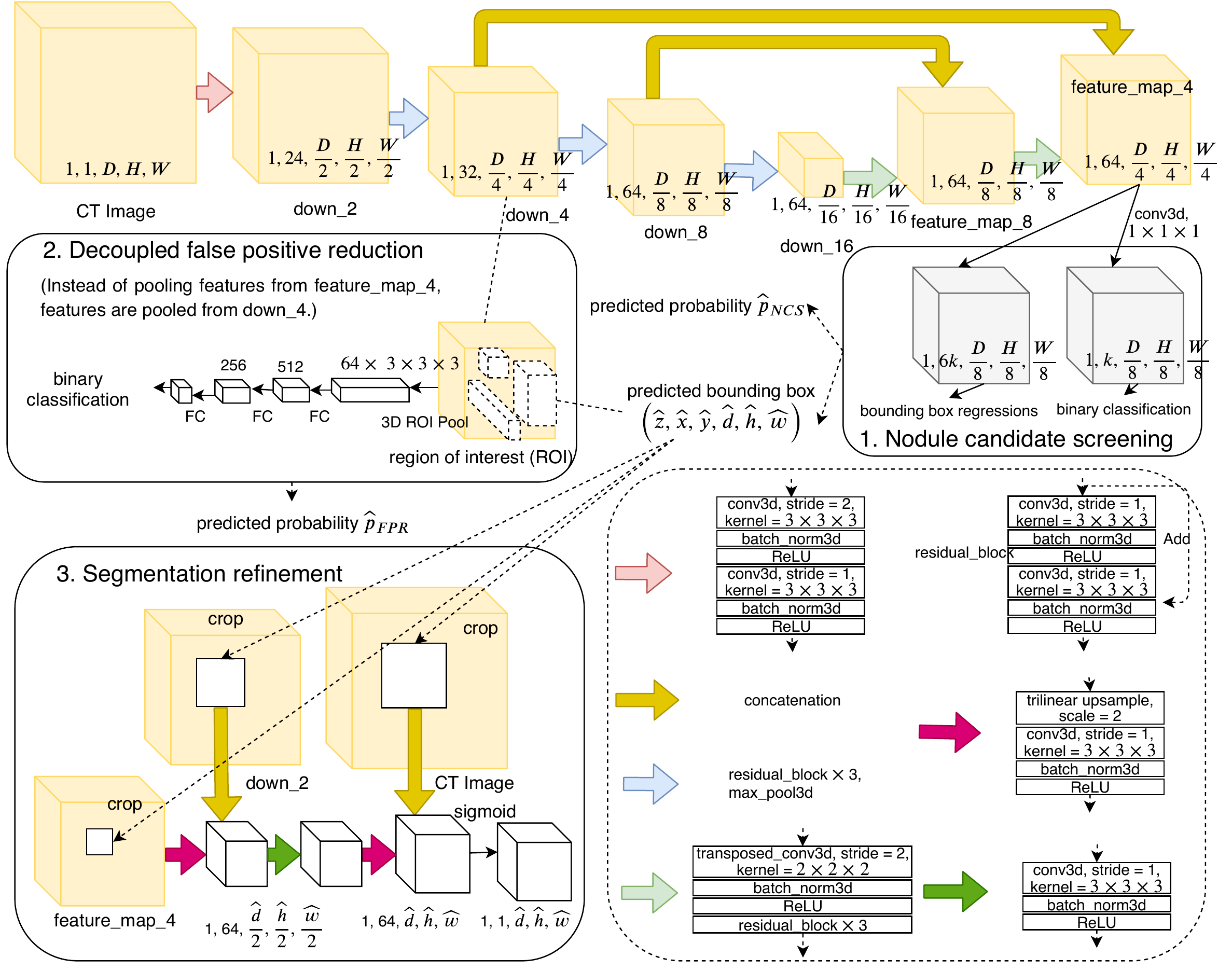}
\caption{\bf Overview of NoduleNet. NoduleNet is an end-to-end framework for pulmonary nodule detection and segmentation, consisting of three sequential stages: nodule candidate screening, false positive reduction and segmentation refinement. $k$ is the number of anchors. FC is short for fully connected layer.}
\label{fig:model}
\end{figure}

Here, we propose a new end-to-end framework, called NoduleNet, for solving pulmonary nodule candidate screening, false positive reduction and segmentation jointly. NoduleNet consists of three parts: nodule candidate screening, false positive reduction and segmentation refinement (\Cref{fig:model}). These three components share the same underlying feature extraction backbone and the whole network is trained in an end-to-end manner. 

% Nodule candidate screening is performed on the feature map that incorporates both high-level image information and low-level voxel information. The results of it are then used to pool image features from the feature map that only has low-level information with small receptive field, to reduce false positives. These two tasks are performed on decoupled feature maps, but share a few early feature maps, allowing learning both task-independent and task-dependent features.

% Finally, the detected bounding boxes are used to crop local image features from all scales of feature maps. Starting from the cropped feature maps with largest scale, the segmentation refinement network progressively upsamples the high-level feature maps and concatenates them with low-level feature maps to generate segmentation mask for each detected pulmonary nodule, at the same scale as the CT image. This approach is different from the mask branch proposed by \cite{he2017mask}, which performs segmentation only using downsampled feature map and then resizes the predicted mask back to the image scale.

Our main contributions are summarized as follows:
\begin{itemize}
    \item We propose a unified model to integrate nodule detection, false positive reduction and nodule segmentation within a single framework, trained end-to-end in a multi-task fashion.
    \item We demonstrate the effectiveness of the model, improving nodule detection accuracy by 10.27\% compared to the baseline model trained only for nodule detection, and achieving a state-of-the-art nodule segmentation accuracy of 83.10\% on Dice-Sørensen coefficient (DSC).
    \item We carry out systematic ablation studies to verify the contributions of several design tricks underlying NoduleNet, including decoupled features maps, segmentation refinement subnet, and multi-task training. 
%     our aforementioned assumptions and intuitions behind the design of each component in NoduleNet, and demonstrate significant improvement on pulmonary nodule detection by using multi-task learning and decoupled feature representations.
% \item  We propose a segmentation refinement branch that generates segmentation mask, using the feature map that has the same scale as the original CT image, and achieve state-of-the-art performance on nodule segmentation with little extra cost.
\end{itemize}
% This approach is different from the mask branch proposed in \cite{he2017mask}. They perform segmentation by only using downsampled feature map and then resize the predicted mask back to the original image scale, which may lose precision due to bounding box regression errors and loss of more fine-grained local features.

% $(b).$ The segmentation refinement network in NoduleNet progressively upsamples the high-level feature maps and concatenates them with low-level feature maps to generate segmentation mask for each detected pulmonary nodule, at the same scale as the CT image. This approach is different from the mask branch proposed by \cite{he2017mask}, that performs segmentation only using downsampled feature map and then resizes the predicted mask back to the image scale.

% Cross validation results on the large-scale LIDC-IDRI \cite{armato2011lung} dataset demonstrate the NoduleNet achieves 10.27\% performance improvement for nodule detection compared to a single stage 3D detector, and an average Dice-Sørensen coefficient (DSC) 83.10\% on nodule segmentation.

% \section{Related work}
% \subsubsection{Nodule detection and false positive reduction}

% \subsubsection{Nodule segmentation}

\section{NoduleNet}
% This section describes in details the architecture of NoduleNet (\Cref{fig:model}). NoduleNet is an end-to-end framework and different networks are trained jointly.

% \subsubsection{Feature extraction backbone}
% The first convolution layer of the feature extraction backbone uses stride two to directly down sample the spatial resolution of the input image volume by half, to save GPU memory. All subsequent feature extraction uses three residual blocks followed by max pooling to reduce spatial resolution by half. A residual block consists of two $3\times 3\times 3$ convolution layers followed by batch normalization and ReLU activation layers. A $1\times 1\times 1$ convolution layer is added if the number of input channels is not equal to the number of output channels. 

% After three residual blocks and three maxpoolings, the feature map has a stride of 16, meaning the resolution of the feature map is $1/16$ of the original image resolution. Then, two upsampling blocks are applied, forming a feature map that has stride 4. Each upsampling block consists of one 3D transposed convolution layer that enlarges the resolution of feature map by a factor of two, and a residual block that takes as input the concatenation of the upsampled feature map and an early feature map that has the same scale.

\subsubsection{Nodule candidate screening (NCS)}
To generate nodule candidates, a $3 \times 3 \times 3$ 3D convolutional layer is applied to the feature map (feature\_map\_4 in Figure~\ref{fig:model}), followed by two parallel $1 \times 1 \times 1$ convolutional layers to generate classification probability and six regression terms associated with each anchor at each voxel on the feature map.
An anchor is a 3D box, which requires six parameters to specify: central z-, y-, x- coordinates, depth, height and width. We chose cube of size 5, 10, 20, 30 and 50 as the 5 anchors in this work. Then, we minimize the same multi-task loss function as \cite{ren2015faster}.

% \begin{equation}
% L(\{p_i\}, \{t_i\}) = \frac{1}{N_{cls}}\sum_i{L_{cls}(p_i, p_i^*)} + \lambda\frac{1}{N_{reg}}\sum_i{L_{reg}(t_i, t_i^*)}
% \label{equation:loss}
% \end{equation}
% where $i$ is the index of $i-th$ anchor in one CT image, $p_i$ is its predicted probability that this anchor contains nodule candidate and $t_i$ is a vector denoting the six parameterized coordinate offsets of this anchor with respect to the ground truth box. $\lambda$ is a hyper parameter balancing the two losses and we set it to 1 in this work. $N_{cls}$ is the total number of anchors chosen for binary classification loss and $N_{reg}$ is the total number of anchors considered for computing regression loss. $p_i^*$ is 0 if $i-th$ anchor does not contain any nodule and 1 otherwise. $t_i^*$ is the ground truth vector for six regression terms and is formally defined as (we ignore subscript $i$ for notational convenience):
% \begin{equation}
% \label{equation:reg_loss}
% \begin{gathered}
% t^*= (t_z, t_y, t_x, t_d, t_h, t_w)\\
% t^*=(\frac{z^*-z_a}{d_a}, \frac{y^*-y_a}{d_a}, \frac{x^*-x_a}{d_a}, \log\frac{h^*}{d_a}, \log\frac{w^*}{h_a}, \log\frac{d^*}{w_a})
% \end{gathered}
% \end{equation}
% $z^*, y^*, x^*, d^*, h^*, w^*$ represent the center coordinates, depth, height and width of the ground truth box. $z_a, y_a, x_a, d_a, h_a, w_a$ denote those for the anchor box. We use weighted binary cross entropy loss with hard negative example mining (OHEM) \cite{shrivastava2016training} and smooth $L1$ loss for $L_{reg}$. The foreground and background ratio of OHEM is set to 1:3 is used in this work.

\subsubsection{Decoupled false positive reduction (DFPR)}
Unlike \cite{ren2015faster} that performs classification using features pooled from the same feature map as RPN (feature\_map\_4). Learning using coupled feature map may lead to sub-optimal solutions of the two tasks. Instead, we use 3D region of interest (ROI) pooling layer to pool features from early feature map that has a small receptive field (down\_4). This not only ensures the false positive reduction network has a small receptive field and can learn feature representations that are substantially different from nodule candidate screening network, but also allows sharing of a few feature extraction blocks. The false positive reduction network minimizes the same multi-task in loss function as the NCS 
% \Cref{equation:loss}, where the $L_{cls}$ is a weighted binary cross entropy loss and $L_{reg}$ remains the same as \Cref{equation:reg_loss}
.

\subsubsection{Segmentation refinement (SR)}
As shown in \Cref{fig:model}, segmentation is performed at the same scale of the original input CT image, by progressively upsampling the cropped high-level feature map (feature\_map\_4) and concatenating them with low-level semantically strong features.

This approach is fundamentally different from the mask branch proposed in \cite{he2017mask}. In \cite{he2017mask}, the authors perform segmentation by only using downsampled feature map and then resize the predicted mask back to the original image scale, which may lose precision due to bounding box regression errors and loss of more fine-grained local features.
% This approach is different from the mask branch proposed by \cite{he2017mask}, which performs segmentation by only using downsampled feature map and then resizes the predicted mask back to the same scale as the original image.

Another advantage is that, only the regions have nodules are upsampled to the original image scale, which only accounts for a small area of the whole input image. This saves a large amount of GPU memory, making whole volume input feasible during training and testing, as compared to upsampling the whole feature map to original input scale in \cite{milletari2016v}. 

The segmentation refinement network minimizes the soft dice loss of the predicted mask sets $\{m\}$ and the ground truth mask sets $\{g\}$ of the input image.
% \begin{equation}
% \label{equation:dice_loss}
% \begin{gathered}
% L_{seg}(\{m\}, \{g\}) = \sum_n^{N_m} 1 - \frac{2\sum_{i=1}^{N_{np}}m_{ni} g_{ni}}{\sum_{i=1}^{N_{np}}m_{ni}g_{ni} + \alpha \sum_{i=1}^{N_{np}}m_{ni}(1-g_{ni}) + \beta \sum_{i=1}^{N_{np}}(1-m_{ni})g_{ni} + \epsilon}
% \end{gathered}
% \end{equation}
% where $N_m$ is the total number of nodules in the input CT scan, $N_{np}$ is the number of pixels in the $n-th$ nodule mask. $m_{ni}$ and $g_{ni}$ denote the predicted probability of the $i-th$ voxel of the $n-th$ mask being a foreground, and the ground truth of that voxel respectively. $\alpha$ and $\beta$ are parameters controlling the trade-off between false positives and false negatives, and we set them both to 0.5 in this work.

\section{Results}
\subsubsection{Data and experiment configurations}
We used LIDC-LDRI \cite{armato2011lung} for evaluting the performance of NoduleNet. LIDC-LDRI is a large-scale public dataset for studying lung cancers, which contains 1018 sets of CT scans collected from multiple sites with various slice thickness. Nodules with diameter equal or greater than 3 mm in this dataset have contour outlined by up to four radiologists. We included only those CT scans met the selection criteria of LUNA16 \cite{setio2017validation} in this work. If the two segmentation masks provided by two radiologists have an intersection over union (IoU) larger than 0.4, we consider the two masks are referring to the same nodule. We consider nodules annotated by at least 3 out of 4 radiologists the ground truth, resulting in a total number of 586 CT scans with 1131 nodules. 
% Note that, although this dataset also includes malignancy of the nodules, in this work, we only focused on the performance of nodule detection and segmentation system. 
Note that the number of CT scans and nodules included in this work may be different from previous work \cite{wang2017central,wu2018joint,aresta2018iw}, due to different inclusion criteria.

A six-fold cross validation was performed to demonstrate the performance of NoduleNet. All models in the experiment were trained using stochastic gradient descent (SGD) with initial learning rate 0.01, momentum 0.9 and $l2$ penalty 0.0001, for 200 epochs. The learning rate was scheduled to decrease to 0.001 after 100 epochs and to 0.0001 after another 60 epochs.

Free-Response Receiver Operating Characteristic (FROC) \cite{kundel2008receiver} analysis was adopted for evaluating the performance of nodule detection. We used the same hit criterium and competition performance metric (CPM) as in the LUNA16 \cite{setio2017validation}. Intersection over union (IoU) and S${\o}$rensen-Dice coefficient (DSC) were used for evaluating the performance of nodule segmentation.

% For evaluating the performance of nodule detection, Free-Response Receiver Operating Characteristic (FROC) \cite{kundel2008receiver} analysis was adopted. FROC analysis was performed by measuring the detection sensitivity and false positives per scan (FPs/scan), where a true positive is defined as the predicted nodule falls within the range $R$ of any ground truth nodule. The final Competition Performance Metric (CPM) is defined as the average sensitivity at seven predefined FPs/scan rates: 1/8, 1/4, 1/2, 1,2, 4, 8.

% Intersection over union (IoU) and Dice-Sørensen coefficient (DSC) were used for evaluating the performance of nodule segmentation.
% IoU and DSC between two masks $M_1$ and $M_2$ are formally defined as:
% \begin{equation}
% \label{equation:iou}
% \begin{gathered}
% \textrm{IoU} = \frac{|M_1 \cap M_2|}{|M_1 \cup M_2|}, \textrm{DSC} = \frac{2 \times |M_1 \cap M_2|}{|M_1| \cup |M_2|}
% \end{gathered}
% \end{equation}

\subsubsection{Nodule detection performance}
In order to fully verify and understand our aforementioned assumptions, we conducted extensive experiments using different network architectures and design choices. We use \textbf{$\textrm{N}_1$} to represent network that only has NCS branch, \textbf{$\textrm{N}_2$} for network has both NCS and FPR branches, and \textbf{$\textrm{N}_3$} for network has all NCS, FPR and SR branches. \textbf{$\textrm{F}_c$} represents the FPR branch is built on the same feature map as NCS, and \textbf{$\textrm{F}_d$} means the FPR branch is built on the decoupled feature map mentioned in previous section. \textbf{R} means the training data is extraly augmented with $xy$ - plane rotation. \textbf{NCS} means the predicted probability comes from NCS branch, \textbf{FPR} means the predicted probability comes from FPR branch, and \textbf{FU} means the predicted probability is fused from NCS and FPR. Note that \textbf{$\textrm{N}_1$} is the widely used 3D RPN for nodule detection \cite{tang2018automated,tang2019end,zhu2017deeplung,liao2019evaluate}, which was served as a strong baseline for evaluating the performance of each added component. The results are summarized in \Cref{table:performance}. 

% A widely used 3D RPN for nodule detection (\textbf{A1}) \cite{tang2018automated,tang2019end,zhu2017deeplung} was served as a strong baseline for evaluating the performance of each component added in the NoduleNet. Note that the nodule candidate screening part of the NoduleNet stand-alone is the widely used 3D RPN. In order to demonstrate the performance gain from adding the segmentation refinement network, we trained NoduleNet both with and without the segmentation network. \textbf{A2} represents NoduleNet trained without mask refienment network, and \textbf{ENS} represents a simple ensemble by averaging the probability of nodule candidate screening network and false positive reduction network. Moreover, to demonstrate the effectiveness of decoupled false positive reduction, we conducted the same experiment as before but using coupled false positive reduction. We use \textbf{V1} to represent models trained using coupled false positive reduction. Lastly, in \textbf{V1} and \textbf{V2}, we only used random jitters, scales for data augmentation while in \textbf{NoduleNet} we also used rotation for data augmentation. The results are summarized in \Cref{table:performance}. 

As seen from \Cref{table:performance}, the sensitivity at 8 false positives per patient rate has a consistent improvement of 1.0\% to 1.5\% by adding the segmentation refinement network (\textbf{$\textrm{N}_3$}), which demonstrates the effectiveness of using the extra nodule segmentation information. 

The average sensitivity of the NoduleNet using decoupled false positive reduction (\textbf{$\textrm{F}_d$}) has around 3\% to 4\% improvement over the NoduleNet using coupled false positive (\textbf{$\textrm{F}_c$}). Moreover, by adding rotation in data augmentation (\textbf{R}), the performance of \textbf{FPR} branch is further improved by around 2.5\% while the performance of \textbf{NCS} branch remains almost the same. This verifies our assumption that classification should learn invariant features, while localization may learn co-variant features. Those findings demonstrate the importance of decoupling modules that are essentially learning different tasks. 

By fusing the predicted probability from NCS and FPR, the performance was consistently improved by 0.7\% - 1.0\%, demonstrating that combining predictions from branches that perceive different level of context information is important.

By adding false positive reduction and segmentation refinement network, the performance of the baseline detector (\textbf{NCS}) is correspondingly improved, showing the effectiveness of multi-task learning and feature sharing. 

All together, NoduleNet outperforms a strong baseline single stage detector by 10.27\%. Note that performance reported in LUNA16 may not be directly comparable to this work, because of different nodule selection criteria, and training and testing data splits. Also, this work focuses on the joint learning of nodule detection and segmentation, whereas the LUNA16 focuses only on nodule detection.

\begin{table}[!ht]
\centering
\begin{tabular}{l c c c c c c c c}
\hline
\textbf{Method} & 0.125 & 0.25 & 0.5 & 1.0 & 2.0 & 4.0 & 8.0 & Avg.\\ \hline
\textbf{\textbf{$\textrm{N}_1$} (\textbf{NCS}) \cite{tang2018automated,tang2019end,zhu2017deeplung,liao2019evaluate}} & 52.17 & 62.51 & 71.09 & 80.46 & 87.27 & 91.07 & 94.43 & 77.00 \\ 
\textbf{\textbf{$\textrm{N}_2$} + \textbf{$\textrm{F}_c$} (\textbf{NCS})} & 53.85 & 62.07 & 71.09 & 79.22 & 86.74 & 90.98 & 93.28 & 76.75 \\ 
\textbf{\textbf{$\textrm{N}_2$} + \textbf{$\textrm{F}_c$} (\textbf{FPR}) \cite{tang2019end}} & 55.79 & 66.93 & 75.77 & 82.40 & 88.68 & 91.78 & 93.10 & 79.21 \\ 
\textbf{\textbf{$\textrm{N}_3$} + \textbf{$\textrm{F}_c$} (\textbf{NCS})} & 53.67 & 63.84 & 74.62 & 83.20 & 88.51 & 92.04 & 94.96 & 78.69 \\ 
\textbf{\textbf{$\textrm{N}_3$} + \textbf{$\textrm{F}_c$} (\textbf{FPR})} & 57.38 & 65.96 & 77.19 & 84.97 & 89.92 & 93.28 & 95.40 & 80.59 \\ \hline
\textbf{\textbf{$\textrm{N}_2$} + \textbf{$\textrm{F}_d$} (\textbf{NCS})} & 56.15 & 66.93 & 74.54 & 82.23 & 88.59 & 92.22 & 95.05 & 79.39 \\ 
\textbf{\textbf{$\textrm{N}_2$} + \textbf{$\textrm{F}_d$} (\textbf{FPR})} & 61.98 & 71.26 & 78.78 & 85.41 & 89.30 & 92.22 & 95.31 & 82.04 \\ 
\textbf{\textbf{$\textrm{N}_3$} + \textbf{$\textrm{F}_d$} (\textbf{NCS})} & 61.45 & 70.20 & 78.16 & 84.62 & 90.27 & 93.63 & 96.20 & 82.08 \\ 
\textbf{\textbf{$\textrm{N}_3$} + \textbf{$\textrm{F}_d$} (\textbf{FPR})} & 68.08 & 73.56 & 81.70 & 85.94 & 90.80 & 93.90 & 96.55 & 84.36 \\ 
\textbf{\textbf{$\textrm{N}_3$} + \textbf{$\textrm{F}_d$} (\textbf{FU})} & 68.70 & 75.60 & 82.23 & 87.36 & 92.04 & 94.96 & 96.46 & 85.34 \\ \hline
\textbf{\textbf{$\textrm{N}_3$} + \textbf{$\textrm{F}_d$} + \textbf{R} (\textbf{NCS})} & 62.78 & 70.65 & 78.43 & 84.44 & 89.74 & 93.10 & 95.49 & 82.09 \\ 
\textbf{\textbf{$\textrm{N}_3$} + \textbf{$\textrm{F}_d$} + \textbf{R} (\textbf{FPR})} & 69.23 & 77.01 & 84.70 & 89.48 & 93.37 & 95.23 & $\textbf{96.55}$ & 86.51 \\ 
\textbf{\textbf{$\textrm{N}_3$} + \textbf{$\textrm{F}_d$} + \textbf{R} (\textbf{FU})} & $\textbf{70.82}$ & $\textbf{78.34}$ & $\textbf{85.68}$ & $\textbf{90.01}$ & $\textbf{94.25}$ & $\textbf{95.49}$ & 96.29 & $\textbf{87.27}$ \\ \hline 
\end{tabular}
\caption{\bf CPM of different methods on the LIDC dataset based on six-fold cross validation. Shown are nodule detection sensitivities (unit: \%) with each column denoting the threshold false positive rate per CT scan (FPs/scan). The last column denotes the average sensitivities across the seven pre-defined FPs/scan thresholds. } 
\label{table:performance}
\end{table}

\subsubsection{Nodule segmentation performance}
\begin{figure}[!h]
\centering
 \includegraphics[width=0.9\textwidth]{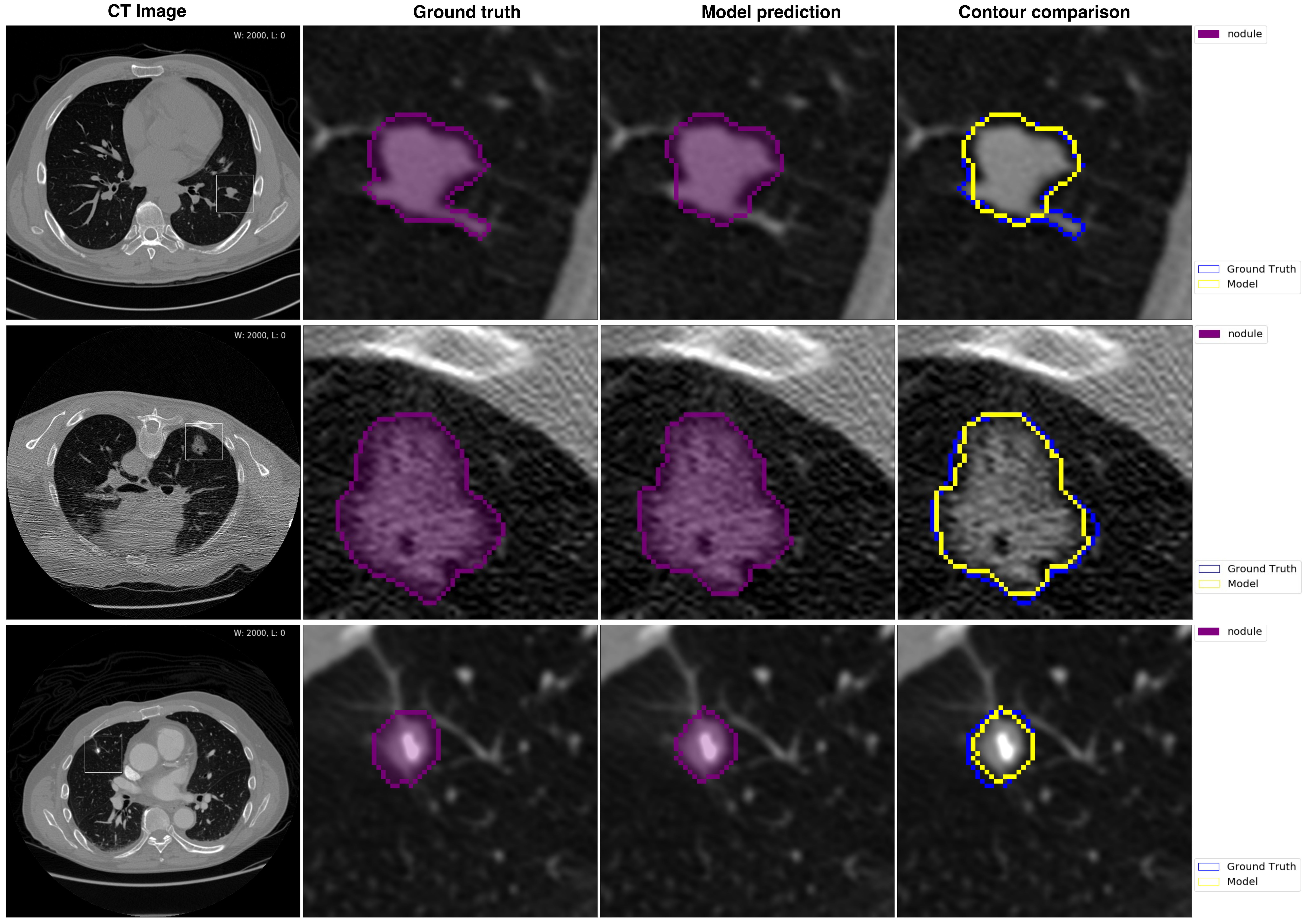}
\caption{\bf Examples of nodule segmentation generated by NoduleNet.}
\label{fig:visualization}
\end{figure}

% Because nodule detection network did not reach a 100\% sensitivity, in order to conduct a fair comparison with state-of-the-art nodule segmentation methods using deep learning, we used the ground truth box as input to the nodule refinement network in this evaluation. 
In \Cref{table:performance_IOU}, we compared the segmentation performance of NoduleNet to other deep learning based methods trained and tested on LIDC dataset \cite{wang2017central,wu2018joint,aresta2018iw}. NoduleNet outperformed previous state-of-the-art deep learning based method by 0.95\% on DSC, without the need to train a separate and dedicated 3D DCNN for nodule segmentation.
% Different methods were trained and tested using different inclusion criteria. 
We randomly selected several nodules for visualizing the segmentation quality (\Cref{fig:visualization}).

\begin{table}[!ht]
\centering
\begin{tabular}{l c c c c c}
\hline
\textbf{Approach} &  \multicolumn{2}{c}{\textbf{\# Nodules}} & \textbf{\# Consensus} & \textbf{IoU} (\%) & \textbf{DSC} (\%)\\
 & \textbf{train} & \textbf{test} & \\ \hline
Wu \etal \cite{wu2018joint} & 1404 & 1404 & 3 & N\textbackslash A & 73.89 $\pm$ 3.87 \\
Aresta \etal \cite{aresta2018iw} & 1593 & 1593 & 3 & 55.00 $\pm$ 14.00 & N\textbackslash A \\
Wang \etal \cite{wang2017central} & 350 & 493 & 4 &71.16 $\pm$ 12.22 & 82.15 $\pm$ 10.76 \\ \hline
NoduleNet & 1131 & 1131 & 3 & 69.98 $\pm$ 10.80 & 81.80 $\pm$ 8.65 \\ 
NoduleNet & 1131 & 712 & 4 & \textbf{71.85 $\pm$ 10.48} & \textbf{83.10 $\pm$ 8.85} \\ \hline
\end{tabular}
\caption{\bf IoU (\%) and DSC (\%) performance of nodule segmentation between different methods. ``\# Consensus'' means each method includes nodules that are annotated by at least ``\# Consensus'' experts.} 
\label{table:performance_IOU}
\end{table}

% \subsubsection{Visualization of nodule segmentation}

\vspace{-0.5cm}
\section{Conclusion}
In this work, we propose a new end-to-end 3D DCNN, named NoduleNet, for solving pulmonary nodule detection, false positive reduction and segmentation jointly. We performed systematic analysis to verify the assumptions and intuitions behind the design of each component in the architecture. Cross validation results on LIDC dataset demonstrate that our model achieves a final CPM score of 87.27\% on nodule detection and DSC score of 83.10\% on nodule segmentation, representing current state-of-the-arts on this dataset. The techniques introduced here are general, and can be readily transferred to other models.   

\bibliographystyle{splncs04}
\bibliography{reference}

\begin{thebibliography}{10}
\providecommand{\url}[1]{\texttt{#1}}
\providecommand{\urlprefix}{URL }
\providecommand{\doi}[1]{https://doi.org/#1}

\bibitem{aresta2018iw}
Aresta, G., Jacobs, C., Ara{\'u}jo, T., Cunha, A., Ramos, I., van Ginneken, B.,
  Campilho, A.: iw-net: an automatic and minimalistic interactive lung nodule
  segmentation deep network. arXiv preprint arXiv:1811.12789  (2018)

\bibitem{armato2011lung}
Armato, S.G., McLennan, G., Bidaut, L., McNitt-Gray, M.F., Meyer, C.R., Reeves,
  A.P., Zhao, B., Aberle, D.R., Henschke, C.I., Hoffman, E.A., et~al.: The lung
  image database consortium (lidc) and image database resource initiative
  (idri): a completed reference database of lung nodules on ct scans. Medical
  physics  \textbf{38}(2),  915--931 (2011)

\bibitem{bray2018global}
Bray, F., Ferlay, J., Soerjomataram, I., Siegel, R.L., Torre, L.A., Jemal, A.:
  Global cancer statistics 2018: Globocan estimates of incidence and mortality
  worldwide for 36 cancers in 185 countries. CA: a cancer journal for
  clinicians  \textbf{68}(6),  394--424 (2018)

\bibitem{cheng2018revisiting}
Cheng, B., Wei, Y., Shi, H., Feris, R., Xiong, J., Huang, T.: Revisiting rcnn:
  On awakening the classification power of faster rcnn. In: Proceedings of the
  European Conference on Computer Vision (ECCV). pp. 453--468 (2018)

\bibitem{ding2017accurate}
Ding, J., Li, A., Hu, Z., Wang, L.: Accurate pulmonary nodule detection in
  computed tomography images using deep convolutional neural networks. In:
  International Conference on Medical Image Computing and Computer-Assisted
  Intervention. pp. 559--567. Springer (2017)

\bibitem{he2017mask}
He, K., Gkioxari, G., Doll{\'a}r, P., Girshick, R.: Mask r-cnn. In: Proceedings
  of the IEEE international conference on computer vision. pp. 2961--2969
  (2017)

\bibitem{kalpathy2016comparison}
Kalpathy-Cramer, J., Zhao, B., Goldgof, D., Gu, Y., Wang, X., Yang, H., Tan,
  Y., Gillies, R., Napel, S.: A comparison of lung nodule segmentation
  algorithms: methods and results from a multi-institutional study. Journal of
  digital imaging  \textbf{29}(4),  476--487 (2016)

\bibitem{khosravan2018s4nd}
Khosravan, N., Bagci, U.: S4nd: Single-shot single-scale lung nodule detection.
  In: International Conference on Medical Image Computing and Computer-Assisted
  Intervention. pp. 794--802. Springer (2018)

\bibitem{kundel2008receiver}
Kundel, H., Berbaum, K., Dorfman, D., Gur, D., Metz, C., Swensson, R.: Receiver
  operating characteristic analysis in medical imaging. ICRU Report
  \textbf{79}(8), ~1 (2008)

\bibitem{liao2019evaluate}
Liao, F., Liang, M., Li, Z., Hu, X., Song, S.: Evaluate the malignancy of
  pulmonary nodules using the 3-d deep leaky noisy-or network. IEEE
  transactions on neural networks and learning systems  (2019)

\bibitem{milletari2016v}
Milletari, F., Navab, N., Ahmadi, S.A.: V-net: Fully convolutional neural
  networks for volumetric medical image segmentation. In: 2016 Fourth
  International Conference on 3D Vision (3DV). pp. 565--571. IEEE (2016)

\bibitem{ren2015faster}
Ren, S., He, K., Girshick, R., Sun, J.: Faster r-cnn: Towards real-time object
  detection with region proposal networks. In: Advances in neural information
  processing systems. pp. 91--99 (2015)

\bibitem{ronneberger2015u}
Ronneberger, O., Fischer, P., Brox, T.: U-net: Convolutional networks for
  biomedical image segmentation. In: International Conference on Medical image
  computing and computer-assisted intervention. pp. 234--241. Springer (2015)

\bibitem{setio2017validation}
Setio, A.A.A., Traverso, A., De~Bel, T., Berens, M.S., van~den Bogaard, C.,
  Cerello, P., Chen, H., Dou, Q., Fantacci, M.E., Geurts, B., et~al.:
  Validation, comparison, and combination of algorithms for automatic detection
  of pulmonary nodules in computed tomography images: the luna16 challenge.
  Medical image analysis  \textbf{42},  1--13 (2017)

\bibitem{tang2018automated}
Tang, H., Kim, D.R., Xie, X.: Automated pulmonary nodule detection using 3d
  deep convolutional neural networks. In: Biomedical Imaging (ISBI 2018), 2018
  IEEE 15th International Symposium on. pp. 523--526. IEEE (2018)

\bibitem{tang2019end}
Tang, H., Liu, X., Xie, X.: An end-to-end framework for integrated pulmonary
  nodule detection and false positive reduction. In: Biomedical Imaging (ISBI
  2019), 2019 IEEE 16th International Symposium on. IEEE (2019)

\bibitem{wang2017central}
Wang, S., Zhou, M., Liu, Z., Liu, Z., Gu, D., Zang, Y., Dong, D., Gevaert, O.,
  Tian, J.: Central focused convolutional neural networks: Developing a
  data-driven model for lung nodule segmentation. Medical image analysis
  \textbf{40},  172--183 (2017)

\bibitem{wu2018joint}
Wu, B., Zhou, Z., Wang, J., Wang, Y.: Joint learning for pulmonary nodule
  segmentation, attributes and malignancy prediction. In: 2018 IEEE 15th
  International Symposium on Biomedical Imaging (ISBI 2018). pp. 1109--1113.
  IEEE (2018)

\bibitem{zhu2017deeplung}
Zhu, W., Liu, C., Fan, W., Xie, X.: Deeplung: 3d deep convolutional nets for
  automated pulmonary nodule detection and classification. arXiv preprint
  arXiv:1709.05538  (2017)

\end{thebibliography}




% LNCS Homepage, \url{http://www.springer.com/lncs}. Last accessed 4
% \end{thebibliography}
\end{document}